\documentclass[letterpaper, 10 pt, conference]{ieeeconf}  

\IEEEoverridecommandlockouts                              

\usepackage{bm, amsmath, amssymb, gensymb, optidef}
\usepackage{algpseudocode, algorithm}
\usepackage{graphicx, subcaption}
\usepackage[switch]{lineno}
\graphicspath{ {images/} }
\usepackage{tikz}

\usepackage{cite}
\makeatletter
\let\NAT@parse\undefined
\makeatother
\usepackage[hyphens]{url}
\usepackage{breakurl}
\usepackage{hyperref}

\title{\LARGE \bf
Geodesic Tracing-Based Kinematic Integration of Rolling and Sliding Contact on Manifold Meshes for Dexterous In-Hand Manipulation
}

\author{Sunyu Wang$^{1}$, Arjun S. Lakshmipathy$^{2}$, Jean Oh$^{1}$, and Nancy S. Pollard$^{1,2}$
\thanks{The authors are with the $^{1}$Robotics Institute and the $^{2}$Computer Science Department at Carnegie Mellon University, Pittsburgh, USA.}
\thanks{Corresponding author's contact: sunyuw@andrew.cmu.edu}
}

\begin{document}
\makeatletter
\let\@oldmaketitle\@maketitle
\renewcommand{\@maketitle}{\@oldmaketitle
\centering
  \includegraphics[width=0.99\linewidth]{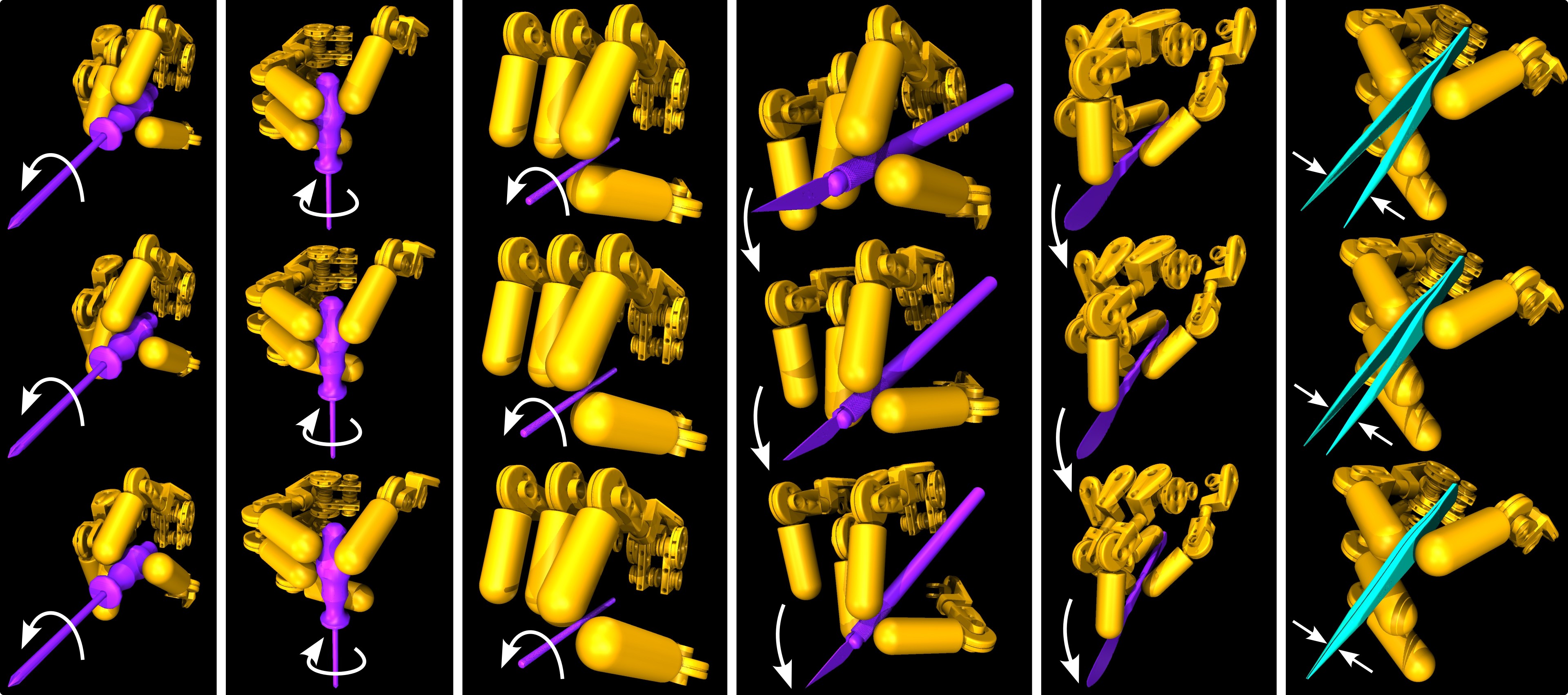}
  \captionof{figure}{Snapshots of a robotic hand performing dexterous in-hand manipulation tasks using the integration scheme we developed. Tasks from left to right: 1) horizontally turning a screwdriver, 2) vertically turning the screwdriver, 3) turning an M2 threaded rod, 4) pressing down a knurled hobby knife, 5) pressing down a dining knife, 6) closing tweezers. }
  \label{fig:headline}
  }
\makeatother
\maketitle
\addtocounter{figure}{-1}

\maketitle
\thispagestyle{empty}
\pagestyle{empty}

\begin{abstract}

Reasoning about rolling and sliding contact, or roll-slide contact for short, is critical for dexterous manipulation tasks that involve intricate geometries. But existing works on roll-slide contact mostly focus on continuous shapes with differentiable parametrizations. This work extends roll-slide contact modeling to manifold meshes. Specifically, we present an integration scheme based on geodesic tracing to first-order time-integrate roll-slide contact directly on meshes, enabling dexterous manipulation to reason over high-fidelity discrete representations of an object's true geometry. Using our method, we planned dexterous motions of a multi-finger robotic hand manipulating five objects in-hand in simulation. The planning was achieved with a least-squares optimizer that strives to maintain the most stable instantaneous grasp by minimizing contact sliding and spinning. Then, we evaluated our method against a baseline using collision detection and a baseline using primitive shapes. The results show that our method performed the best in accuracy and precision, even for coarse meshes. We conclude with a future work discussion on incorporating multiple contacts and contact forces to achieve accurate and robust mesh-based surface contact modeling. 

\end{abstract}
\section{Introduction}

Societal expectations have been growing for humanoid robots to contribute to manufacturing, assembly, agriculture, and healthcare \cite{humanoid_news_1, humanoid_news_2, humanoid_news_3}. To meet such expectations, humanoid robots need to have strong dexterous manipulation capabilities, which remain challenging to achieve largely because modeling and reasoning about contact is difficult. 

A seminal work on contact modeling is the rigid-body single-point roll-slide contact kinematics model formulated by Montana in 1988 \cite{montana_1988}. This model considers the contacting bodies' detailed geometries, addresses both rolling and sliding motions, and has seen iterations of improvements and different formulations \cite{canny_1990, cui_2009, cui_2010, cui_2015, cui_2017, bulut_2018, cui_textbook_2020, ugurlu_2019, ugurlu_2020}. Meanwhile, significant research progress has been made on dexterous manipulation motion planning with roll-slide contact \cite{li_1989, cole_1989, trinkle_1989, montana_1995, chen_2000, jeong_2013, bai_2014, noohi_2015, nakashima_2017, endo_2019, cortez_2023, zong_2023, tang_2024, livnat_2024, yang_2025}. 

Yet, related prior works mostly focused on primitive shapes with globally differentiable parametric functions, whereas many real-world objects cannot be easily modeled in these approaches. Furthermore, decomposing a complex shape into primitive shapes introduces artificial discontinuities, making it difficult to model continuous global contact motion. 

In contrast, manifold meshes are performant and versatile in discretely representing realistic shapes---they are the data structure of numerous CAD models and 3D digital art works, the blueprints for computer-aided manufacturing processes such as 3D printing and CNC machining. Hence, extending roll-slide contact modeling to manifold meshes can enable dexterous manipulation to reason over high-fidelity discrete representations of an object's true geometry beyond continuous shapes. We believe that this has the potential to unlock new capabilities for multi-finger robotic hands. 

To this end, we present a geodesic tracing-based integration scheme to first-order time-integrate roll-slide contact directly on manifold meshes. We used our method to plan motions of a multi-finger robotic hand performing dexterous in-hand manipulation tasks, as shown in Fig. \ref{fig:headline}. The planning was achieved with a least-squares optimizer that computes the most stable instantaneous grasp by minimizing contact sliding and spinning. Then, we evaluated our method against two baselines---one using collision detection and one using primitive shapes---in two simulation experiments: a hand turning a screwdriver, and a sphere rolling on a ring under different mesh resolutions. The results show that our method performed the best in accuracy and precision, even for coarse meshes, with limitations in handling multiple contacts. We conclude with a discussion on resolving these limitations while incorporating contact forces to create an accurate and robust mesh-based surface contact model. 

\section{Related Works}

\subsection{Roll-Slide Contact Modeling}

The roll-slide contact kinematics model we refer to in this work is a set of first-order differential equations that characterizes the positions and velocities of contact points on two rigid surfaces that roll and slide on each other. Some oldest versions of this model date back to the 1980s \cite{cai_1987, montana_1988}. Since then, this subject has been extensively studied in rigid body kinematics \cite{canny_1990, cui_2009, cui_2010, cui_2015, cui_2017}, and the results are summarized in robotics textbooks \cite{mason_1985, mls_1994, mason_2001, lynch_2017, cui_textbook_2020}. Meanwhile, improvements and alternative formulations have been proposed \cite{bulut_2018, ugurlu_2019, ugurlu_2020}. A key difference among these formulations is their surface parametrizations. 

However, these prior works typically concern continuous geometries with global, differentiable parametrizations. For geometries that are difficult to parametrize with closed-form differentiable functions, the methods in these prior works may not be readily applicable. Our work builds on these prior works' ideas---particularly Montana's formulation \cite{montana_1988}---but focuses on the \textit{discrete} setting with manifold meshes, which do not always have closed-form global parametrizations.

\subsection{Dexterous Motion Planning with Roll-Slide Contact}

The roll-slide contact kinematics model has been linked to dexterous manipulation motion planning for multi-finger robotic hands since its inception. Early works on this subject absorbed the roll-slide contact model into the robotic hand's kinematics and dynamics equations \cite{li_1989, cole_1989, trinkle_1989, montana_1995, chen_2000}. Through Jacobians, contact type models, grasp matrix, and inverse kinematics and dynamics, dexterous motions were generated. However, as existing roll-slide contact models mostly rely on differentiable surface parametrizations, these early works only considered primitive shapes as the manipulated objects. 

Later, trajectory optimization and policy learning approaches surged in popularity in dexterous manipulation motion planning. As a more classical method, the roll-slide contact model has often acted as a foundational but supporting part in a larger algorithmic framework. Several works did exploit rolling motions in hand motion planning, but few addressed roll-slide contact kinematics for detailed geometries of general objects \cite{jeong_2013, bai_2014, noohi_2015, nakashima_2017, endo_2019, cortez_2023, zong_2023, tang_2024, livnat_2024}. 

A closely related work to ours is \cite{yang_2025}, which employed signed distance field to parametrize general shapes in a trajectory optimization framework that emphasizes differentiability. The authors generated dexterous motions for a multi-finger robotic hand, and impressively demonstrated them in the real world. However, a limitation is that \cite{yang_2025} still modeled the manipulated objects as primitive shapes, with the authors themselves noting that accepting meshes as input is generally intractable. Our method offers a solution to this limitation by taking full manifold meshes as input. 

\section{Methods}


Our method uses Montatna's roll-slide contact kinematics model \cite{montana_1988} as the basis. Hence, we will first recapitulate Montana's theory, and then present our method on top. 

\subsection{Induced Velocity of Contact Frame}
Consider a general 3D rigid body with a body frame $\{ B \}$ fixed to it. Assume that the body has an orthogonal coordinate chart $f: \mathbb{R}^2 \mapsto \mathbb{R}^3$ that describes its surface relative to $\{ B \}$. Define a contact frame $\{ C \}$, whose origin moves on the surface and whose $z$-axis is parallel with the surface normal at $\{ C \}$'s origin. Define $g \in \mathbb{R}^2$ as the minimal-coordinate position of $\{ C \}$'s origin relative to $\{ B \}$ through $f$, and $\dot{g}$ as its time derivative. Then, $\{ C \}$'s generalized velocity relative to $\{ B \}$ expressed in $\{ C \}$, or $\{ C \}$'s body twist, is
\begin{align}
    V^{b}_{BC} = \begin{bmatrix} \omega^{b}_{BC} \\ v^{b}_{BC} \end{bmatrix} = \begin{bmatrix} E K M \\ TM \\ M \\ 0_{1 \times 2} \end{bmatrix} \dot{g}, 
    \text{where } E = \begin{bmatrix} 0 & -1 \\ 1 & 0 \end{bmatrix}. 
    \label{eq:induced_contact_velocity}
\end{align}

$\omega^{b}_{BC}, v^{b}_{BC} \in \mathbb{R}^3$ are $\{ C \}$'s angular and linear velocities relative to $\{ B \}$ expressed in $\{ C \}$, respectively. $M \in \mathbb{R}^{2 \times 2}$ is the metric tensor, $K \in \mathbb{R}^{2 \times 2}$ is the curvature tensor, and $T \in \mathbb{R}^{1 \times 2}$ is the torsion form. Next, define a local frame $\{ L \}$ that instantaneously coincides with $\{ C \}$ but is \emph{fixed} to the body. This means that $\{ L \}$ has the same pose, i.e., position and orientation, as $\{ C \}$, but zero velocities relative to $\{ B \}$. 


\subsection{Roll-Slide Contact Kinematics}
Now, consider two rigid bodies indexed $0$ and $1$. With the above definitions, we have six frames: $\{ B_{i} \}, \{ C_{i} \}, \{ L_{i} \}, i \in \{ 0, 1 \}$. Without loss of generality, define each body's surface normal to be outward-pointing. Then, an ideal point contact between the two bodies means that their contact frames' origins coincide and $z$-axes anti-align. If the contact is maintained, the relative velocities between $\{ C_{0} \}$ and $\{ C_{1} \}$ will be zero except the angular element about the $z$-axis, which is the spin element. Mathematically, 
\begin{align}
    T_{ C_{0} C_{1} } = \begin{bmatrix}
        * & * & 0 & 0 \\ * & * & 0 & 0 \\ 0 & 0 & -1 & 0 \\ 0 & 0 & 0 & 1
    \end{bmatrix}, 
    V^{b}_{C_{0} C_{1}} = \begin{bmatrix} 0_{2 \times 1} \\ * \\ 0_{3 \times 1} \end{bmatrix}.
    \label{eq:ideal_contact}
\end{align}
$T_{ C_{0} C_{1} }$ is the homogeneous transformation representing $\{ C_{1} \}$'s pose relative to $\{ C_{0} \}$. A ``$*$" means any scalar. We refer to (\ref{eq:ideal_contact}) as the ideal point contact constraint. 

Since $\{ L_{i} \}$ coincides with $\{ C_{i} \}$ but is fixed to body $i$, the relative twist between $\{ L_{0} \}$ and $\{ L_{1} \}$, $V^{b}_{L_{0} L_{1}} = \begin{bmatrix}
    \omega_{x} & \omega_{y} & \omega_{z} & v_{x} & v_{y} & v_{z}
\end{bmatrix}^{\intercal}$, represents the Cartesian rolling and sliding velocities between the two bodies. $\omega_{z}$ is the spinning speed about the contact normal, $v_{z}$ is the separation/interpenetration speed, and $v_{x}, v_{y}$ are the sliding speeds in the contact tangent plane. By twist frame transformation \cite{mls_1994}, $V^{b}_{L_{0} L_{1}}$ and $V^{b}_{C_{0} C_{1}}$ are related by
\begin{align}
    V^{b}_{L_{0} L_{1}} + V^{b}_{B_{1} C_{1}} = \text{Ad}_{T_{C_{1} C_{0}}} V^{b}_{B_{0} C_{0}} + V^{b}_{C_{0} C_{1}},
    \label{eq:roll_slide_kin}
\end{align}
where $\text{Ad}$ is the adjoint. (\ref{eq:roll_slide_kin}) and (\ref{eq:induced_contact_velocity}) connect the Cartesian rolling and sliding velocities and the contacts' minimal-coordinate velocities, $\dot{g}_{i}$'s, of both bodies, forming the governing equations of roll-slide contact kinematics. 

\subsection{Body-to-Contact Integration with Collision Detection}
After reviewing Montana's theory, we investigate first-order time integration of roll-slide contact on manifold meshes with the target application of dexterous manipulation motion planning. Since what matters is the relative motion between two contacting bodies, without loss of generality, we define body 0 as the reference body whose pose and body twist relative to a stationary world frame $\{ W \}$ are known, and body 1 as the moving body. Then, the problem is to time-update the poses of $\{ C_{0} \}, \{ C_{1} \}$, and $\{ B_{1} \}$ relative to $\{ W \}$, given both bodies' poses, their contacts' poses, and the relative contact twist, $V^{b}_{L_{0} L_{1}}$, at the current time step.  

An intuitive solution is to integrate the bodies' velocities first, and then update the contacts based on closest points. Applying twist frame transformation \cite{mls_1994} to $V^{b}_{L_{0} L_{1}}$ yields
\begin{align}
    V^{b}_{L_{0} L_{1}} = \text{Ad}_{T_{L_{1} B_{1}}} V^{b}_{W B_{1}} - \text{Ad}_{T_{L_{1} B_{0}}} V^{b}_{W B_{0}}, 
    \label{eq:relative_body_twists}
\end{align}
where $V^{b}_{W B_{i}}$ is $\{ B_{i} \}$'s body twist relative to $\{ W \}$. Since $T_{W B_{0}}, T_{W B_{1}}, T_{B_{0} L_{0}}, T_{B_{1} L_{1}}, V^{b}_{W B_{0}}$, and $V^{b}_{L_{0} L_{1}}$ are known, we can solve (\ref{eq:relative_body_twists}) for $V^{b}_{W B_{1}}$ and use any standard integrator to integrate the body twists and update the poses. 

Next, we update the contact frames. If the two bodies interpenetrate, we update each body's contact point as the point on that body closest to the penetration depth-weighted average of the interpenetration points; the deeper the penetration, the larger the weight. If the two bodies separate, we update each body's contact point as the closest point on that body to the other. Then, we update the contact frames' $z$-axes as the linearly interpolated vertex normals at the updated contact points. The $x$- and $y$-axes can be updated arbitrarily as long as they form valid rotation matrices with the $z$-axes. For ease of visualization, we update the $x$-axes as the vectors pointing from the current contacts to the updated contacts, orthogonally projected onto the $z$-axes. Lastly, we update the $y$-axes with cross product and normalize all three axes to unit vectors. We implemented this method as a baseline using a collision detector \cite{trimesh, python_fcl, fcl}, a closest point querier \cite{fcpw}, and the 4th-order Runge-Kutta (RK4) integrator in MuJoCo \cite{mujoco}, with all dynamics---including gravity---turned off. 

\subsection{Contact Stabilization}
Despite its intuitiveness, the collision detection-based baseline cannot always keep the two bodies in an ideal point contact. This is because explicit time integration of velocities can produce drift and motions in the contact tangent plane, causing interpenetration or separation. 


To resolve this issue, we implemented an inverse kinematics-inspired contact stabilizer \cite{lynch_2017}, shown in Algorithm \ref{alg:contact_stabilizer}. At each time step, the stabilizer acts as a reactive feedback controller and generates a stabilizing twist $V_{stbl}$ that, if added to $V^{b}_{L_{0} L_{1}}$, pushes the current $T_{C_{0} C_{1}}$ closer to the ideal $T_{C_{0} C_{1}}$ in (\ref{eq:ideal_contact}). In Algorithm \ref{alg:contact_stabilizer}, $T_{C_{0} C_{1}}$ is the two contact frames' current relative pose. $T_{C_{0} C_{1}, ideal}$ is the $T_{C_{0} C_{1}}$ in (\ref{eq:ideal_contact}). $\omega_{stbl}, v_{stbl}$ are the angular, linear velocities in $V_{stbl}$. $\text{expm} \left( \cdot \right)$ and $\text{logm}\left( \cdot \right)$ are matrix exponential and matrix logarithm. 
\begin{algorithm}
\caption{Contact Stabilizer}\label{alg:contact_stabilizer}
\begin{algorithmic}
\State $T_{error} \leftarrow T_{C_{0} C_{1}}^{-1} T_{C_{0} C_{1}, ideal}$
\State $T_{error} \left[:3, :3 \right] \leftarrow \text{expm} \left( a_{z_{error}} \right)$
\State $V_{stbl} \leftarrow \text{logm} \left( T_{error} \right)$
\State $V^{b}_{L_{0} L_{1}} \leftarrow V^{b}_{L_{0} L_{1}} + \begin{bmatrix} k_{\omega} \omega_{stbl}^{\intercal} & k_{v} v_{stbl}^{\intercal} \end{bmatrix}^{\intercal}$
\end{algorithmic}
\end{algorithm}

Unlike in the inverse kinematics case, the two contact frames' $x$- and $y$-axes do not need to coincide. Hence, before taking the matrix logarithm, we replace the rotation matrix in $T_{error}$ in Algorithm \ref{alg:contact_stabilizer} with a rotation matrix computed from $a_{z_{error}}$, an axis-angle vector that anti-aligns only the two contact frames' $z$-axes. Lastly, two independently tuned gains, $k_{\omega}$ and $k_{v}$, are applied to the angular and linear velocities in $V_{stbl}$, before it is added to $V^{b}_{L_{0} L_{1}}$. We implemented this contact stabilizer for the collision detection-based baseline. 

\subsection{Contact-to-Body Integration with Geodesic Tracing}
Another approach to time-integrating roll-slide contact on meshes is to apply Montana's theory. However, a mesh is generally non-smooth and does not always have a differentiable parametrization in a global coordinate chart. Also, our target application is dexterous manipulation motion planning. It is inadequate to define a local coordinate chart based on a prescribed differentiable space curve as the contact's path. 


Our solution is to use the contact frame itself as the coordinate chart that locally parametrizes the surface in the Euclidean space, and update the contact frame at every time step. Hence, this local coordinate chart only needs to stay valid for one time step. Specifically, we define the minimal-coordinate position $g$ in (\ref{eq:induced_contact_velocity}) as the signed geodesic distances from the contact frame's origin in the positive $x$- and $y$-directions. To time-integrate the contact point, we trace the geodesic defined by $g$ relative to the body frame $\{ B \}$ using explicit Euler and a mesh-based geodesic tracer \cite{geometry_central}. This ensures that the integrated contact point stays on the mesh. After the integration, we update the contact frame with the linearly interpolated vertex normals as its $z$-axis. The other two axes can be arbitrary as long as they form a valid rotation matrix with the $z$-axis, and we update them in the same way as in the collision detection-based baseline. Algorithm \ref{alg:contact_integrator_geodesic} shows this procedure with time step size $\Delta t$. 

\begin{algorithm}
    \caption{Geodesic Tracing-Based Contact Integrator}\label{alg:contact_integrator_geodesic}
    \begin{algorithmic}
    \State $\dot{g}_{0}, \dot{g}_{1} \leftarrow \text{roll-slide kinematics} \left( V^{b}_{ L_{0} L_{1} }, K_{0}, K_{1} \right)$ (\ref{eq:induced_contact_velocity}), (\ref{eq:roll_slide_kin})
    \State $p_{B_{i} C_{i}} \leftarrow \text{trace geodesic} \left( \dot{g}_{i} \Delta t \right), i \in \{ 0, 1 \}$ \cite{geometry_central}
    \State $R_{B_{i} C_{i}} \leftarrow \text{update contact frame} \left( p_{B_{i} C_{i}} \right)$
    \end{algorithmic}
\end{algorithm}
$R_{B_{i} C_{i}}$ and $p_{B_{i} C_{i}}$ are $\{ C_{i} \}$'s rotation matrix and Cartesian position relative to $\{ B_{i} \}$, respectively. Based on $g_{i}$'s definition, the metric tensor $M$ in (\ref{eq:induced_contact_velocity}) is the identity matrix. $K_{i}$ is the curvature tensor at the current contact point on body $i$, which is estimated using a central difference based on geodesic tracing and linearly interpolated vertex normals. Since the $x$- and $y$-axes are not tracked, we ignore the torsion form and the spin element in (\ref{eq:induced_contact_velocity}). 

After updating $R_{B_{i} C_{i}}$ and $p_{B_{i} C_{i}}$, since the reference body's updated pose is known, a rigid body transformation would yield the moving body's pose that exactly satisfies the ideal point contact constraint (\ref{eq:ideal_contact})
\begin{align}
    T_{W B_{1}} = T_{W B_{0}} T_{B_{0} C_{0}} T_{C_{0} C_{1}} \left( \psi \right)  T_{B_{1} C_{1}}^{-1}. 
    \label{eq:exact_mating}
\end{align}
We refer to (\ref{eq:exact_mating}) as exact mating. However, since spin is not tracked, the top-left $2 \times 2$ block of $T_{C_{0} C_{1}}$ is unknown, as in (\ref{eq:ideal_contact}). This means that the moving body has spin as an unconstrained degree of freedom (DoF), which we parametrize with a scalar $\psi$. To resolve this issue, we compute a ``shadow orientation" of the moving body, $R'_{W B_{1}}$, by integrating its angular velocity with RK4. Though $R'_{W B_{1}}$ may not satisfy the contact constraint, it helps determine $\psi$, which we solve for by minimizing the axis-angle vector error between $R'_{W B_{1}}$ and $R_{W B_{1}}$ with iterative inverse kinematics \cite{lynch_2017}. 

The method above ensures that the integrated frames exactly satisfy the ideal point contact constraint (\ref{eq:ideal_contact}). Yet, for closed kinematic chains, such as a multi-finger grasp, additional structural constraints are present, which may render the entire kinematic system overdetermined. In those cases, we use the contact stabilizer in Algorithm \ref{alg:contact_stabilizer} instead. 

\subsection{Hand Motion Planning with Roll-Slide Contact}
Consider a multi-finger hand in contact with an object. The kinematic objective of dexterous in-hand manipulation is to achieve desired poses and twists of the object via the hand's movement. Hence, we define the object as the reference body with a known initial pose and twist trajectory. 

To plan the hand's motion, we let the hand maintain the most stable instantaneous grasp. Kinematically, this means to minimize the sliding and spinning speeds at each contact while preventing separation and interpenetration, which translates into a constrained least-squares optimization
\begin{align}
    \dot{u}_{H} = \text{argmin} \left( \omega_{z}^2 + v_{x}^{2} + v_{y}^{2} \right) \text{ subject to } v_{z} = 0.
    \label{eq:min_sliding}
\end{align}
$\omega_{z}$ and $v_{x}, v_{y}, v_{z}$ are the spin and the linear elements of $V^{b}_{L_{0} L_{1}}$, respectively. $\dot{u}_{H}$ is the hand's minimal-coordinate velocities, usually finger joint and palm velocities. To apply (\ref{eq:min_sliding}) to a hand holding an object, we employ (\ref{eq:roll_slide_kin}) and Jacobians to express $V^{b}_{L_{0} L_{1}}$ in terms of minimal-coordinate velocities: $V^{b}_{L_{0} L_{1}} = J_{H} \dot{u}_{H} + J_{O} \dot{u}_{O}$, where $J_{H}$ and $J_{O}$ are the hand's and the object's Jacobians, respectively, and $\dot{u}_{O}$ is the object's minimal-coordinate velocity, which is known. Substituting this equation into (\ref{eq:min_sliding}) yields $\dot{u}_{H}$. We implemented this procedure with an off-the-shelf constrained least-squares optimizer \cite{qpsolvers, clarabel_2024}. In addition, we added two regularization terms to (\ref{eq:min_sliding}), one to encourage finger movement and discourage palm movement, and the other to smooth the trajectory of $\dot{u}_{H}$. 



\begin{figure*}[t]
    \centering
    \includegraphics[width=0.95 \textwidth]{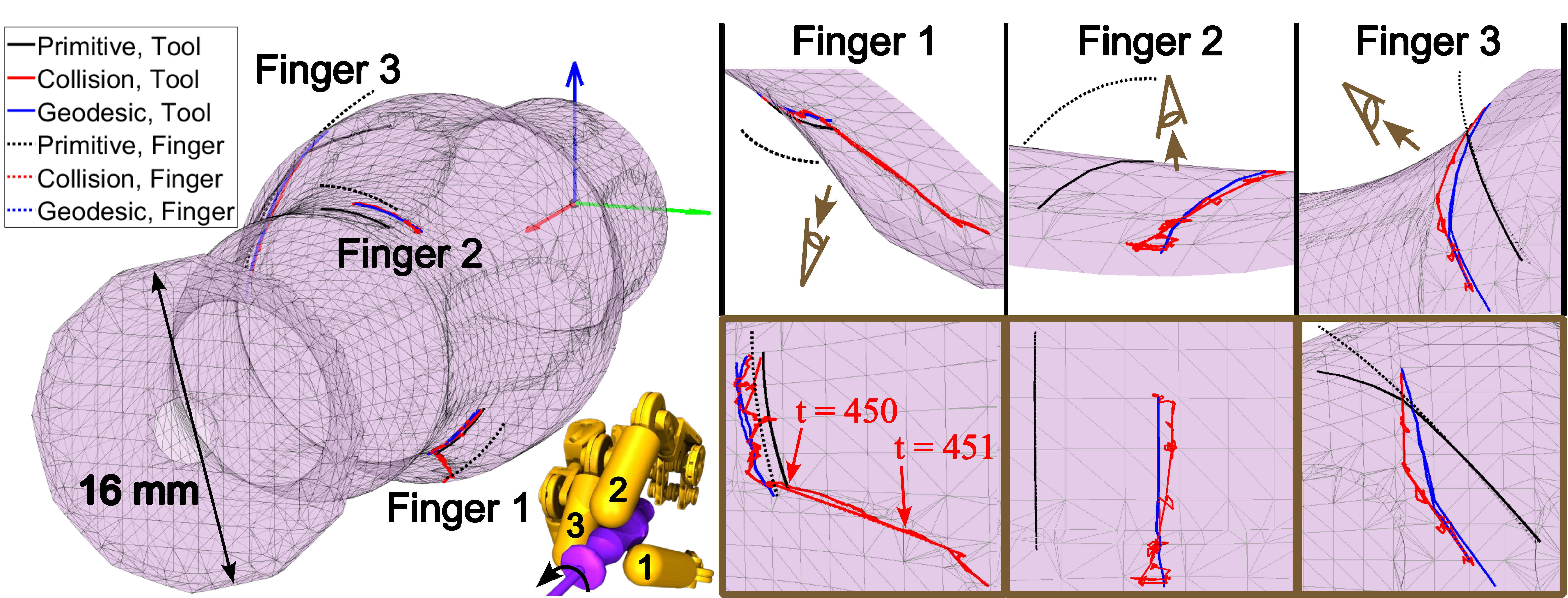}
    \caption{Contact paths on the screwdriver's handle relative to the screwdriver's body frame. On the right, each column contains a detail view and a normal view of each finger's contact paths. The brown eye symbol and arrow indicate the incident light direction in which the normal view in the corresponding bottom row in brown box is captured. The red arrows point at finger 1's contact points produced by the collision baseline at time steps $450$ and $451$, respectively. The ``Tool" and ``Finger" suffixes in the legend mean that the corresponding contact path is on the screwdriver or the finger. }
    \label{fig:paths_screwdriver}
\end{figure*}

\begin{figure}[t]
    \centering
    \includegraphics[width=0.45 \textwidth]{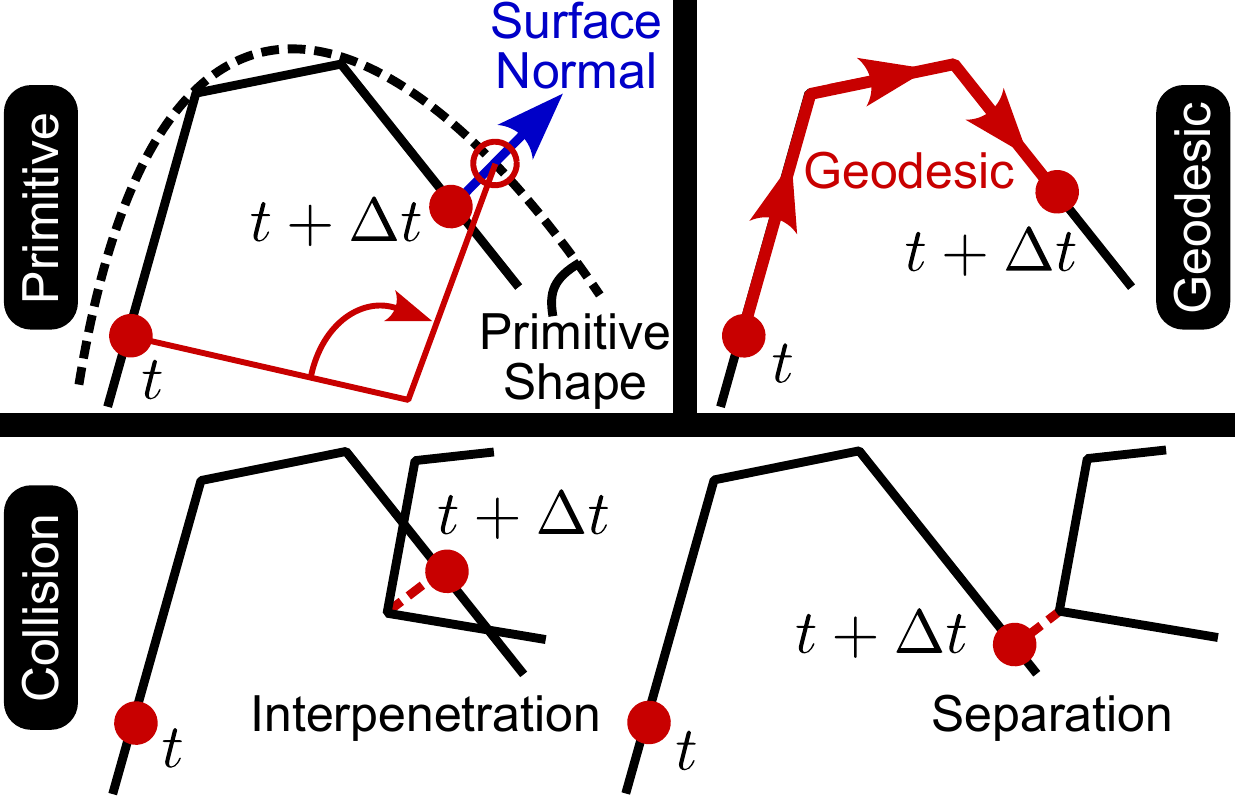}
    \caption{One-time step integration from the same contact point at time $t$ with the primitive baseline (top left), the geodesic method (top right), and the collision baseline (bottom) on a 3-edge 2D mesh. The black solid lines are the mesh. The red points are the contact points. }
    \label{fig:methods_comparison}
\end{figure}

\begin{figure}[t]
    \centering
    \includegraphics[width=0.48 \textwidth]{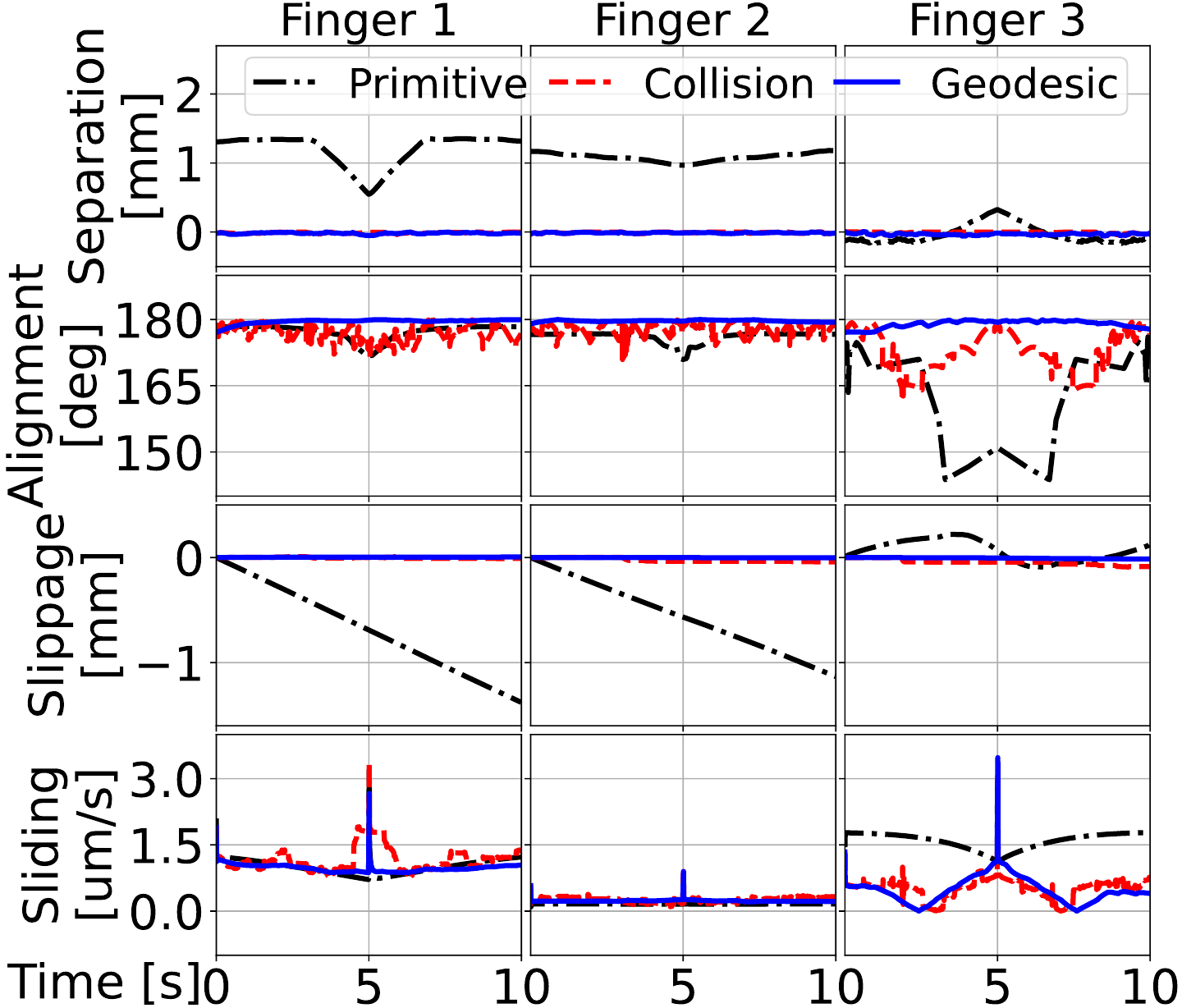}
    \caption{Screwdriver experiment metrics. Each column represents a finger. The subplots on each row have the same vertical axis ticks. ``um" means micrometer. }
    \label{fig:results_screwdriver}
\end{figure}

\section{Results and Discussion}
\subsection{Hand Motion Planning with Meshes from the Internet}
We downloaded five objects' meshes from online \cite{grab_cad} and imported them to the simulation for hand motion planning. The objects are shown in Fig. \ref{fig:headline}. The screwdriver has concavities for grip, the threaded rod has threads, and the hobby knife's handle is knurled---all are intricate geometric features difficult to model with primitive shapes or differentiable parametrizations. The tweezers have one revolute joint for each arm at its base. The hand consists of four identical 3-DoF fingers, indexed 1--4 from the thumb to the ring finger. Each finger has a capsule-shape distal link, can flex/extend and adduct/abduct at the MCP joint, and flex/extend at the IP joint. The palm is modeled as a 6-DoF floating base. 

To plan the hand's motion, we first manually configured the hand into the pose for holding each object. Then, we prescribed each object's twist as a constant, and planned the finger joint and palm velocities with (\ref{eq:min_sliding}) at every time step. Each object moved in one direction for $5$ seconds, and immediately moved in the opposite direction with the same speed for another $5$ seconds. The simulation ran at $100$ Hz. Fig. \ref{fig:headline} shows the motion snapshots in the first $5$ seconds.


Overall, the hand motions are realistic, especially the twisting motions for the screwdriver and the threaded rod, which engage the fingers' adduction/abduction DoF. However, the collision detection-based baseline produced noisy contact movements, whereas the geodesic tracing-based method produced interpenetration between the fingers and the threaded rod. 

\subsection{Quantitative Experiment: Hand Turning Screwdriver}
To quantitatively compare these methods, we computed four metrics for each contact during the horizontal screwdriver motion shown in the first column of Fig. \ref{fig:headline}. The metrics are explained below. The ideal values mean the metrics for an ideal point contact with zero sliding. 
\begin{enumerate}
    \item Separation, computed as the mean of all interpenetration points' depths if two meshes interpenetrate, or the minimum distance between two meshes if they separate. A positive, negative value means separation, interpenetration, respectively. Ideal value $= 0$ mm. 
    \item Alignment, computed as the angle between the contact frames' $z$-axes on two meshes. Ideal value $= 180 \degree$. 
    \item Slippage, computed in three steps: a) record the geodesic distance traveled by the contact on each mesh at every time step, b) sum these distances from start time to current time, c) subtract the sums between the two meshes. Ideal value $= 0$ mm. 
    \item Sliding, computed as the sliding speed $\sqrt{v_{x}^{2} + v_{y}^{2}}$ from the least-squares (\ref{eq:min_sliding}). Ideal value $= 0$ mm/s. 
\end{enumerate}

Additionally, we added a primitive shape-based baseline by fitting an ellipsoid to the screwdriver's handle and a cylinder and a hemisphere to the finger's distal link. Then, we computed the contacts' minimal coordinate velocities with (\ref{eq:induced_contact_velocity}) and 
cylindrical- and spherical-coordinate parametrizations. Lastly, we integrated these velocities using explicit Euler without contact stabilization as in Algorithm \ref{alg:contact_stabilizer} or exact mating as in (\ref{eq:exact_mating}). Since the primitive shapes are abstract models and the meshes are the true geometries, we computed the metrics for this baseline by projecting the contact point on the primitive shape onto its corresponding mesh using ray tracing. Fig \ref{fig:methods_comparison} compares the primitive shape-based baseline with the collision detection-based baseline and the geodesic tracing-based method. For conciseness, we abbreviate the three methods' names to ``the primitive baseline", ``the collision baseline", and ``the geodesic method". 

\begin{figure}[t]
    \centering
    \includegraphics[width=0.48 \textwidth]{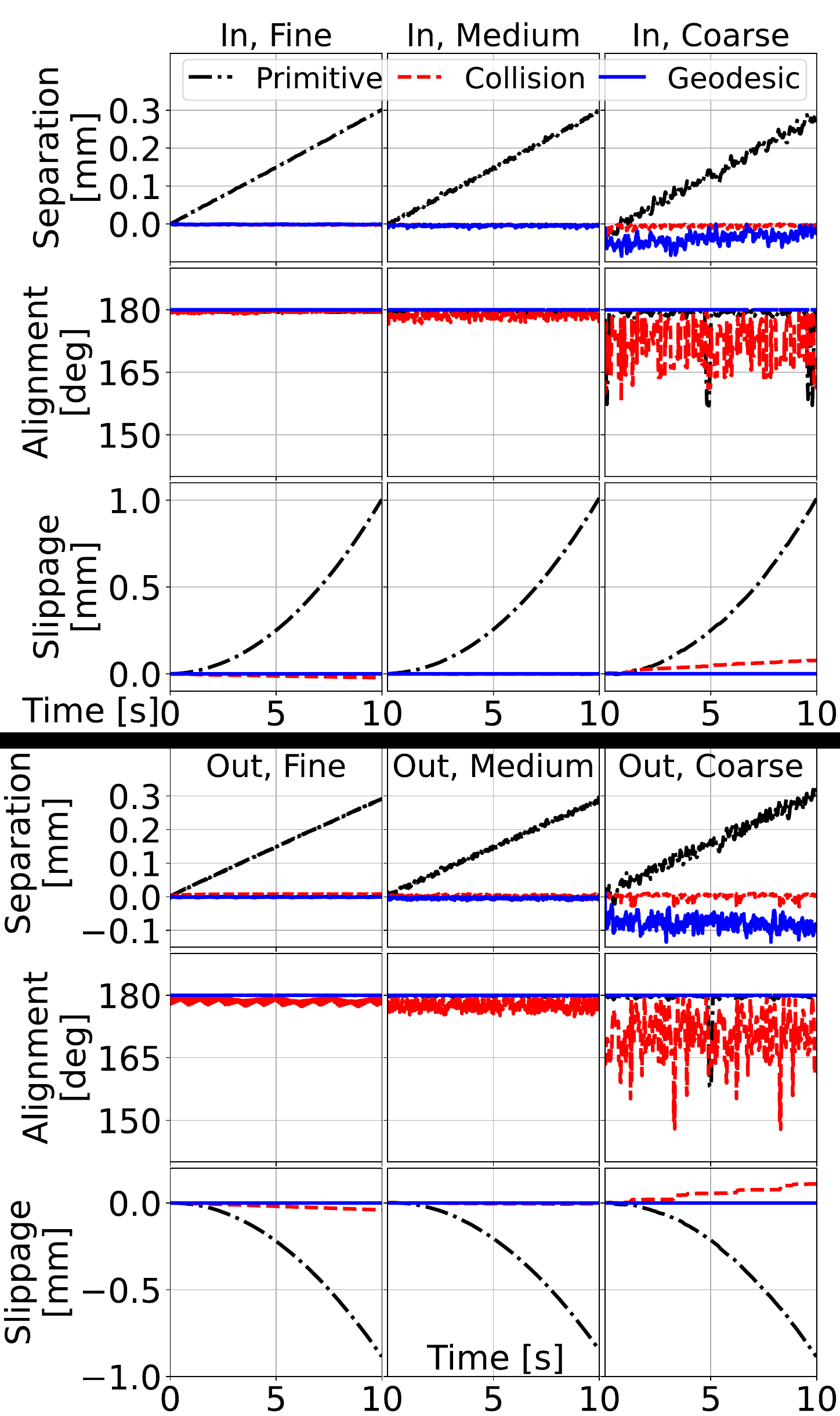}
    \caption{Sphere experiment metrics. The top and bottom $9$ plots with ``In" and ``Out" in the titles are for the sphere rolling inside and outside the ring, respectively. Each column represents a mesh resolution. }
    \label{fig:results_sphere}
\end{figure}

\begin{figure}[t]
    \centering
    \includegraphics[width=0.48 \textwidth]{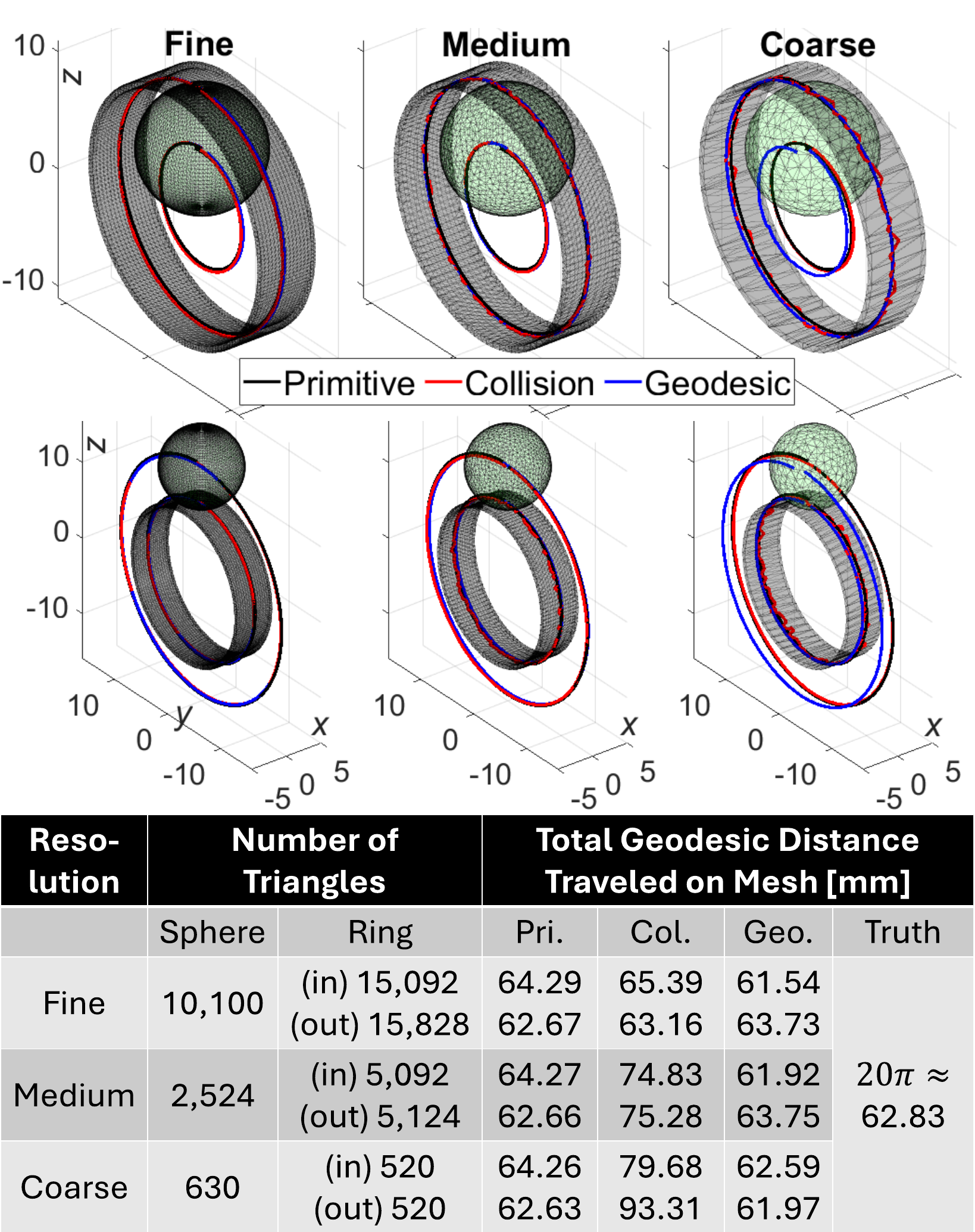}
    \caption{(Top) Sphere experiment paths. Each row represents a side of the ring. Each column represents a mesh resolution. The paths on the rings are the contacts' paths. The paths not on the rings are the paths of the sphere's centroid. The length unit is millimeters. (Bottom) Sphere experiment tabular data. Pri., Col., Geo., and Truth stand for the primitive baseline, the collision baseline, the geodesic method, and the ground truth, respectively. }
    \label{fig:paths_sphere}
\end{figure}

Fig. \ref{fig:paths_screwdriver} and \ref{fig:results_screwdriver} show the results. We make the following observations: 1) The fingers maintained contact with the screwdriver for the collision baseline and the geodesic method, but lost contact for the primitive baseline, as shown by the ``Separation" metric in Fig. \ref{fig:results_screwdriver}. Meanwhile in Fig. \ref{fig:paths_screwdriver}, if a finger is in contact with the screwdriver, the contact path on the screwdriver and the contact path on the finger should coincide, which is not the case for the primitive baseline, indicating separation. This is expected, as an ellipsoid cannot accurately fit the screwdriver's detailed geometries, and nothing in the primitive baseline prevents the explicit integration drift. 2) Both the collision baseline and the geodesic method produced metrics close to the ideal values. But in Fig. \ref{fig:results_screwdriver}, the ``Alignment" and ``Sliding" trajectories for the geodesic method are smoother. Also, in Fig. \ref{fig:paths_screwdriver}'s bottom row, the geodesic method's contact paths resemble smooth versions of the collision baseline's contact paths in parts where the two methods' paths overlap. Both phenomena indicate that the geodesic method is more precise. This is reasonable, as the geodesic method traces the incremental geodesics integrated from the geodesic velocities exactly on the meshes, preserving a sense of spatial continuity for the contact frames. In contrast, the collision baseline incorporates only positions without considering velocities. Consequently, a contact point updated by the collision baseline can be anywhere on the mesh. 3) Shown in Fig. \ref{fig:paths_screwdriver}'s bottom row, from time step $t = 450$ to $t = 451$, finger 1's contact point produced by the collision baseline traveled for a considerably long geodesic distance. When this happened, finger 1's ``Separation", ``Alignment", and ``Slippage" metrics remained close to their ideal values. Only the collision baseline exhibited this behavior. We interpret this as a discontinuous contact shift in a situation where more than one possible contact point exist due to the geometry around the contact. This behavior occurred only for the collision baseline because it does not preserve spatial continuity and considers all possible contact points on the entire mesh. In contrast, the geodesic method obeys the point contact model and integrates velocities to positions, making it unlikely to discontinuously break a contact and establish a distinct new contact. This also explains why the collision baseline is less prone to interpenetration than the geodesic method for the threaded rod. 4) In Fig. \ref{fig:results_screwdriver}, the sliding speeds are small relative to the screwdriver's size, indicating that the fingers mostly rolled on the screwdriver. The spikes at the $5$th second are due to the abrupt reversal of the screwdriver's prescribed velocity.

\subsection{Quantitative Experiment: Sphere Rolling on Ring}
We compared the three methods in a simpler experiment of a sphere rolling on the inside and outside of two rings. The side of the ring on which the sphere rolls always has a diameter of $20$ mm. The sphere's diameter is $10$ mm. Both the sphere and the rings have fine, medium, and coarse mesh resolutions. The simulation ran at $100$ Hz for $10$ seconds. The sphere's angular velocity was prescribed so that the sphere would travel for exactly one revolution around the ring in $10$ seconds if the shapes were continuous. The sphere purely rolls, meaning that the contact sliding speed is always zero. Hence, we changed the ``Sliding" metric from the screwdriver experiment to the total geodesic distance traveled, which is averaged between the contacts on the sphere and the ring. This metric's ground truth is the ring's circumference under continuous geometry: $20 \pi$. Since the kinematic system is an open chain, we used the exact mater in (\ref{eq:exact_mating}) instead of the contact stabilizer in Algorithm \ref{alg:contact_stabilizer} for the geodesic method. 

Fig. \ref{fig:results_sphere} and \ref{fig:paths_sphere} show the results. Our observations are as follows: 1) Without contact stabilization or exact mating, the primitive baseline drifted quickly and could not maintain the contact. Though these drifts are not obvious on the contact paths in Fig. \ref{fig:paths_sphere}, they are significantly more severe than the results from the other two methods, as reflected by the ``Separation" metric in Fig. \ref{fig:results_sphere}. 2) The collision baseline's metrics and contact paths became noisier, and its total geodesic distance traveled drastically deviated from the ground truth as mesh resolution decreased. This is because, for the collision baseline, the contact traveled increasingly laterally on the ring for coarser meshes, generating the zig-zag contact paths in Fig. \ref{fig:paths_sphere}. 3) The geodesic method almost exactly maintained the pure rolling motion and the ideal point contact for fine and medium mesh resolutions. For coarse meshes, interpenetration became larger and the sphere's traveling direction was off from the ring's center, causing a slightly spiral path of the sphere's centroid. This is because the meshes developed flatter faces and sharper corners as their resolutions decreased, whereas their vertex normals were still linearly interpolated and the exact mater (\ref{eq:exact_mating}) was used. When the meshes were too coarse, the interpolation could no longer obtain surface normals close to those on a smooth sphere or ring. Thus, interpenetration and movement direction drift became more likely. Nevertheless, the geodesic method's total geodesic distance traveled remained close to the ground truth for coarse meshes, as shown in Fig. \ref{fig:paths_sphere}. 

\subsection{Limitations and Future Work}
The geodesic method has limitations in handling multiple contact points. As the screwdriver experiment showed, multi-contact situations are hard to avoid for complex shapes. Hence, a meaningful future direction would be to extend roll-slide contact modeling to surface contact between meshes. A promising approach to achieving this is to combine the geodesic method and the collision baseline, as they appear complementary. The geodesic method excels at accuracy and precision, but is local. The collision baseline excels at recognizing all contacts on the entire mesh, but lacks precision and robustness against mesh resolution degradation. Therefore, the two methods can be combined into a hierarchical integration scheme: The collision baseline globally detects the most likely contacts in low precision, and the geodesic method finely integrates each contact. 


Another limitation is that this work concerns first-order kinematic integration, which involves poses and velocities but not forces. This means that the planned motions are kinematically feasible, but may not be feasible when gravity, friction, and inertial forces are present. Despite this, considering the complex kinematic and contact constraints in dexterous multi-finger in-hand manipulation, the planned motions contain detailed information about roll-slide behaviors, and could be close to well-conditioned trajectories feasible in the real world, which could be useful for downstream trajectory optimization, policy learning, and animation tasks. 

Meanwhile, we envision the incorporation of soft contact models, such as pressure field contact \cite{hydroelastic, pfc_velocity}, to produce the center of pressure as a representative contact point. This could enable a point contact model to describe the essential movement trend of a surface contact patch for motion planning, which is an exciting direction to explore. 

\section{Conclusions}
This work presents a geodesic tracing-based method to first-order time-integrate roll-slide contact directly on manifold meshes. The significance is that our method enables roll-slide contact modeling and dexterous manipulation motion planning to reason over high-fidelity discrete representations of an object's true geometry, beyond primitive shapes and shapes with differentiable parametrizations. Our method's technical core is to use the contact frame as the coordinate chart to locally parametrize the surface in the Euclidean space, and apply Montana's theory with a mesh-based geodesic tracer. We used our method to plan the motion of a multi-finger robotic hand performing dexterous in-hand manipulation tasks. The results are qualitatively realistic. Then, we quantitatively evaluated our method in two simulation experiments against a baseline using collision detection and a baseline using primitive shapes. The results show that our method performed the best in accuracy, precision, and robustness against low mesh resolution, with limitations in handling multiple contacts. Lastly, we discuss future directions to complementarily combine our geodesic method and the collision baseline, and to incorporate contact forces via soft contact models and the center of pressure. 
\section*{Acknowledgment}
This research was partially supported by NSF Convergence Accelerator award ITE-2344109, by the National Institute of Biomedical Imaging and Bioengineering of the National Institutes of Health under Award Number R01EB036842, and by the RAI Institute.


\bibliographystyle{IEEEtran}
\bibliography{humanoids_2025.bib}


\end{document}